\documentclass[runningheads]{llncs}
\usepackage{graphicx}
\usepackage{amssymb}
\usepackage{amsmath}
\usepackage{threeparttable}
\usepackage[marginal]{footmisc}
\usepackage{threeparttable}
\usepackage{multirow}
\usepackage{graphicx}
\begin{document}
\title{PFA-ScanNet: Pyramidal Feature Aggregation with Synergistic Learning
	for Breast\\ Cancer Metastasis Analysis}
\titlerunning{PFA-ScanNet for Breast Cancer Metastasis Analysis}
%
\author{Zixu Zhao\inst{1}\and
	Huangjing Lin\inst{1} \and
	Hao Chen\inst{2} \and
	Pheng-Ann Heng\inst{1,3}}
\authorrunning{Z. Zhao et al.}

\institute{Department of Computer Science and Engineering,\\
	The Chinese University of Hong Kong, Hong Kong SAR, China \and
	Imsight Medical Technology, Co., Ltd. Hong Kong SAR, China \and
	Guangdong Provincial Key Laboratory of Computer Vision and Virtual Reality Technology, Shenzhen Institutes of Advanced Technology, Chinese Academy of Sciences, Shenzhen, China}

\maketitle              

\begin{abstract}
Automatic detection of cancer metastasis from whole slide images (WSIs) is a crucial step for following patient staging and prognosis. Recent convolutional neural network based approaches are struggling with the trade-off between accuracy and computational efficiency due to the difficulty in processing large-scale gigapixel WSIs. To meet this challenge, we propose a novel Pyramidal Feature Aggregation ScanNet (PFA-ScanNet) for robust and fast analysis of breast cancer metastasis. Our method mainly benefits from the aggregation of extracted local-to-global features with diverse receptive fields, as well as the proposed synergistic learning for training the main detector and extra decoder with semantic guidance. Furthermore, a high-efficiency inference mechanism is designed with dense pooling layers, which allows dense and fast scanning for gigapixel WSI analysis. As a result, the proposed PFA-ScanNet achieved the state-of-the-art FROC of 90.2\% on the Camelyon16 dataset, as well as competitive kappa score of 0.905 on the Camelyon17 leaderboard. In addition, our method shows leading speed advantage over other methods, about 7.2 min per WSI with a single GPU, making automatic analysis of breast cancer metastasis more applicable in the clinical usage.
\end{abstract}

\section{Introduction}

The prognosis of breast cancer mainly focuses on grading the stage of cancer, which is measured by the tumor, node, and distant metastasis (TNM) staging system \cite{sobin1997tnm}. 
With the boosting progress in high-throughput scanning and artificial intelligence technology, automatic detection of breast cancer metastasis in sentinel lymph nodes has great potential in cancer staging to assist clinical management. 
The algorithm is expected to detect the presence of metastases in five slides with lymphatic tissues dissected from a patient, and measure their extent to four metastasis categories and finally grade the pathologic N stage (pN-stage) following the TNM staging system. However, the task is challenging due to several factors: (1) the difficulty in handling large-scale gigapixel images (e.g., 1-3 GB per slide); (2) the existence of hard mimics between normal and cancerous region; (3) the significant size variance among different metastasis categories.

Recently, many deep learning based methods adopt patch-based models to directly analyze whole slide images (WSIs) \cite{lunit,liu2017detecting,li2018cancer,wang2016deep}. The most common way is to extract small patches in a sliding window manner and feed them to the model for inference. For example, ResNet-101 and Inception-v3 are leveraged as the backbone of detectors in \cite{lunit} and \cite{liu2017detecting}, bringing the detection results to 85.5\% and 88.5\% with regard to FROC on Camelyon16 dataset, respectively. However, the patch-based inference leads to dramatically increased computational costs when applied to gigapixel WSI analysis, which are not applicable in clinical usage. To reduce the computational burden, Kong et al. utilized a lightweight network (student network) supervised by a large capacity network (teacher network) with transfer learning \cite{kongbin}. Also, Lin et al. proposed a modified fully convolutional network (FCN), namely Fast ScanNet, to overcome the speed bottleneck by allowing dense scanning in anchor layers \cite{fastscannet}. These scan-based models reduce redundant computations of overlaps for faster inference but are hampered by limited discrimination capabilities. Encoding multi-scale features is still beyond attainment for scan-based models due to their relatively simple network architectures.

Another challenging problem of lymph node classification lies in how to effectively retrieve tiny metastasis, i.e., ITC ($ <$ 0.2mm) and micro-metastasis ($ <$ 2mm), while rejecting most of the hard mimics.
Several methods \cite{wang2016deep,fastscannet} circumvent false positives via hard negative mining, which focus on the most challenging negative patches. This seems to benefit the performance overall but decreases the sensitivity on small ITC lesions remarkably. 
Furthermore, it may disintegrate the prediction into pieces due to mimic patches existed in metastatic regions, leading to inaccurate evaluation on metastasis size. 
To tackle this issue, Li et al. proposed a neural conditional random field (NCRF) deep learning framework \cite{li2018cancer} combining with hard negative mining. Although spatial correlations is considered, it still achieved limited performance on metastasis detection. Other type of guidance, e.g., semantic guidance that helps the model distinguish hard mimics, has never been incorporated into the detections methods for cancer metastasis.

Aiming at developing a detection system as accurate as possible while maintaining the efficiency, we propose a novel Pyramidal Feature Aggregation ScanNet (PFA-ScanNet). Our contributions are threefold:
(1) We raise a novel way to aggregate pyramidal features for scan-based model which can increase its discrimination capability. Specifically, we focus on local-to-global features extracted from pyramidal features by proposed Parameter-efficient Feature Extraction (PFE) modules.
(2) A high-efficiency inference mechanism is designed with dense pooling layers, allowing the detector to take large sized images as input for inference while being trained in a flexible patch-based fashion.
(3) Synergistic learning is proposed to collaboratively train the detector and decoder with semantic guidance, which can improve the model's ability to retrieve metastasis with significantly different size.  
Overall, our method achieved the highest FROC (90.2\%), competitive AUC (99.2\%) and kappa score (0.905) on Camelyon16 and Camelyon17 datasets with faster inference speed.

\begin{figure}
	\centering
	\includegraphics[width=\textwidth]{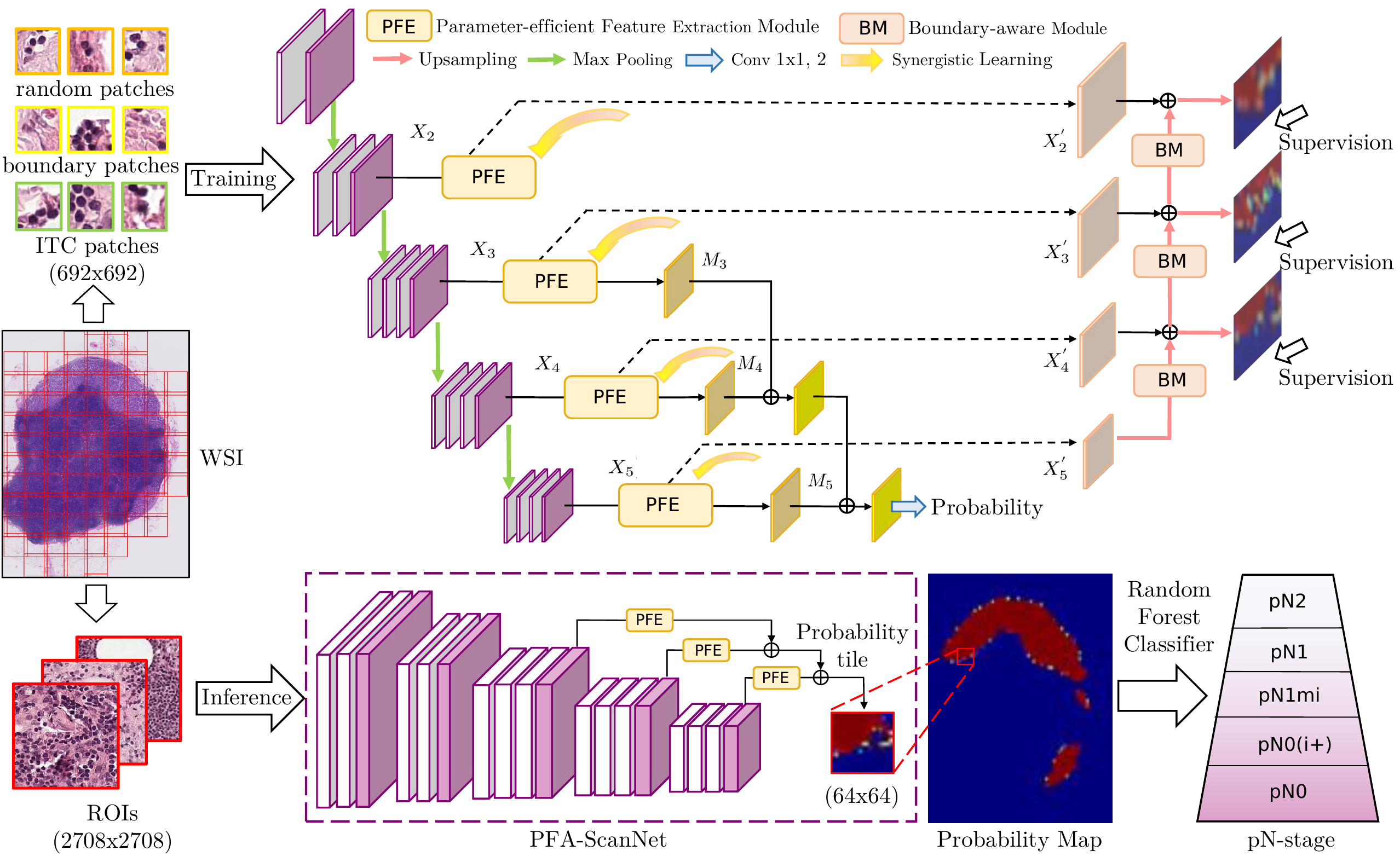}
	\caption{An overview of the proposed PFA-ScanNet }
	\label{fig:net}
\end{figure}
\section{Method}

The proposed PFA-ScanNet is a scan-based fully convolutional network consisting of a main detector for classification and an extra decoder for segmentation. As shown in Fig. \ref{fig:net}, Parameter-efficient Feature Extraction (PFE) modules are integrated into the detector at each feature level to extract local-to-global features with diverse receptive fields and less parameters. The extracted features are then aggregated in a top-down path (detector) and a bottom-up path (decoder). 
\subsection{Pyramidal Feature Aggregation for Accurate Classification }
\begin{figure}
	\centering
	\includegraphics[width=\textwidth]{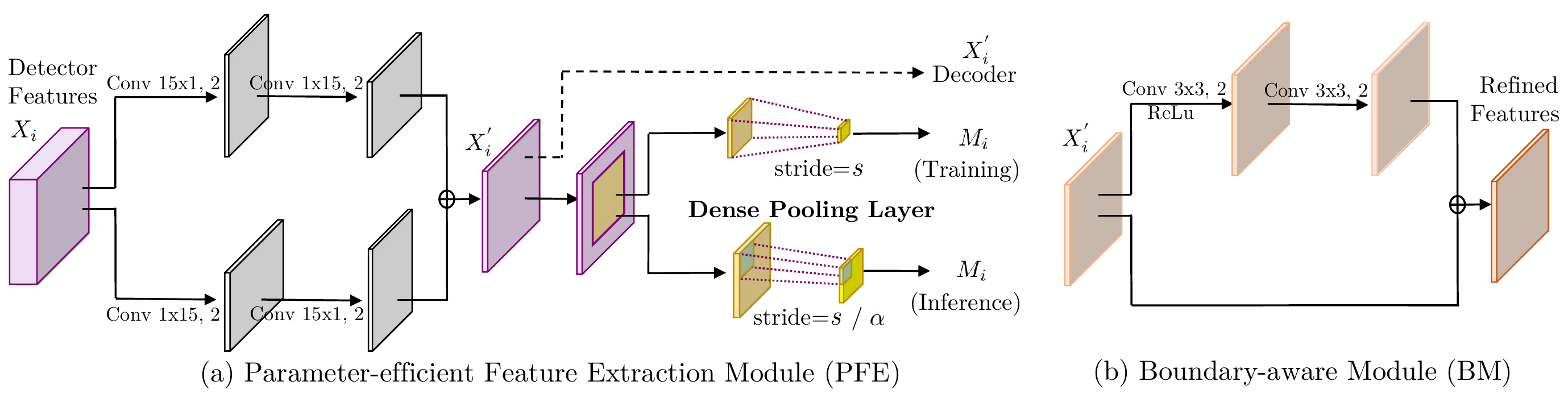}
	\caption{Detail structure of (a) Parameter-efficient Feature Extraction Module (PFE) and (b) Boundary-aware Module (BM) in proposed PFA-ScanNet.}
	\label{fig:module}
\end{figure}

Inspired by the feature pyramid network \cite{lin2017feature}, we propose to make full use of pyramidal features for a scan-based model. We firstly raise the Parameter-efficient Feature Extraction (PFE) module to extract local-to-global features with less parameters from pyramidal features. It also benefits the fast inference in section 2.2. Fig. \ref{fig:module}(a) shows the detailed structure of PFE. Let $ \left\{X_{i} \right\} $ denotes the pyramidal feature generated by the detector at feature level $ i$ $(i =2,3,4,5)$. $ X_{i} $ is firstly passed through a global convolution \cite{Peng_2017_CVPR} with a large kernel to enlarge the receptive fields and reduce the feature map number. 
To further reduce the computation burden and number of parameters, we employ symmetric and separable large filters, which is a combination of $ 1\times15+15\times1 $ and $ 15\times1+1\times15 $ convolutions instead of directly using larger kernel of size $ 15\times15 $. 
To formulate local-to-global features from the refined feature $ X_{i}^{'} $, regions with $\left\{1/4, 1/2, 1/1 \right\}$ size of $ \left\{ X_{3}^{'}, X_{4}^{'}, X_{5}^{'} \right\}$ are cropped to capture diverse receptive fields before average pooling layer. The local-to-global features $ M_{i} $ is then added with $ M_{i+1} $ from a higher feature level and finally passed through a $ 1\times 1 $ convolution with softmax activation to predict the probability. In this way, our detector can efficiently encode pyramidal features and present a strong discrimination capability. 
\subsection{WSI Processing with High-efficiency Inference}
To meet the speed requirement, we inherit the no-padding FCN in Fast ScanNet \cite{fastscannet} as the trunk of our detector but remove the last three fully convolutional layers where the computation is expensive in the inference phase. Unlike the anchor layer raised in \cite{fastscannet}, we propose a dense pooling layer in PFE which allows dense scanning with little extra cost. A dense coefficient $ \alpha $ is introduced in the dense pooling layer to control the pooling strides of average pooling operation. The pooling strides $  \left\{128, 64, 32 \right\} $ are associated with feature level $ \left\{3,4,5\right\} $ in the training phase and will be converted to $ \left\{128/\alpha, 64/\alpha, 32/\alpha \right\} $ in the inference phase. It allows dense and fast predictions when $ \alpha$ increases in the form of $ \alpha = 2^{n} \times 16$ $(n=0,1,2,...)$. Accordingly, our network can take regions of interest (ROIs) with a size of $ L_{R} $ as input for inference while being trained with small patches with a size of $ L_{p} $ for extensive augmentation. In other words, our network inherently falls into the category of FCN architecture, which is equivalent to a patch-based CNN with input size $ L_{p} $ and scanning stride $ S_{p} $, but the inference speed becomes much faster by removing redundant computations of overlaps. To better understand this mechanism, we denote the scanning stride for refetching ROIs as $ S_{R} $ and size of the predicted probability tile as $ L_{m} $, and summarize the rules for high-efficiency inference as follows: 
\begin{equation}
\begin{cases} 
L_{R} = L_{p} + (L_{m}-1) \times (S_{p} / \alpha), \\
S_{R} = (S_{p} / \alpha) \times L_{m} ,\\
\end{cases}
\end{equation}
\subsection{ Semantic Guidance with Synergistic Learning}
Given that the surrounding tissue region is helpful to determine whether the small patch is metastasis or not, we develop our network with an extra decoder branch to synergistically learn the semantic information along with the detector. In the decoder, feature map $ X_{i}^{'} $ generated in PFE is firstly passed through a Boundary-aware Module (BM) to refine the boundary of the metastatic region. As shown in Fig. \ref{fig:module}(b), BM models the boundary alignment in a residual structure \cite{Peng_2017_CVPR} to take advantage of the local contextual information and localization cue. Afterwards, the generated feature is upsampled with deconvolutions and then added with $ X_{i-1}^{'} $ of higher resolutions to generate new score maps in a bottom-up path. Deep supervision is injected to specific layers to learn the multi-level semantic information, which can also speed up the convergence rate.

The basic idea of synergistic learning is training the detector and decoder simultaneously. Nevertheless, it is hard to minimize the classification loss and segmentation loss simultaneously in one iteration. This is caused by the mislabelled region and zigzag boundaries existed in WSI annotations, which have the tendency to overwhelm other informative regions in segmentation loss calculation and thus dominate the gradients. To solve the problem, we modify the binary cross-entropy loss into a truncated form \cite{zhou2019cia} that can reduce the contribution of outliers with high confidence prediction. Our segmentation loss is shown as follows: 
\begin{equation}
\mathcal{L}_{seg}(\mathcal{X};\mathcal{W})=
\begin{cases} 
\sum\limits_{x\in\mathcal{X}}\sum\limits_{t\in\left\{0, 1\right\} }  (-\log(\gamma)+ \frac{1}{2}(1-\frac{p^{2} (t|x; \mathcal{W})}{\gamma^{2}})),p (t|x;\mathcal{W})<\gamma\\
\sum\limits_{x\in\mathcal{X}}\sum\limits_{t\in\left\{0, 1\right\} }-\log(p (t|x; \mathcal{W}))  \qquad \qquad \quad  ,p (t|x; \mathcal{W})\geqslant\gamma \\
\end{cases}
\label{loss}
\end{equation}
where $\mathcal{W}  $ denotes parameters of our model, $ \mathcal{X} $ denotes the training patches, and $ p (t|x; \mathcal{W})$ is the predicted probability for the ground truth label $ t $ given the input pixel $ x $. The segmentation loss will clip outliers at the truncated point $ \gamma \in[0, 0.5] $ when $  p (t|x; \mathcal{W})<\gamma$, while preserving the loss value for others. Therefore, it can ease the gradient domination and benefit the learning of informative regions. When $ \gamma=0 $, it will degrade into binary cross-entropy.
Meanwhile, we directly employ the binary cross-entropy loss as our classification loss to train the detector. Let $ \mathcal{W}_{d} $ denote parameters in the detector and $ \lambda $ be the trade-off hyperparameter, the overall loss function for synergistic learning is defined as:
\begin{equation}
\mathcal{L}_{total}(\mathcal{X};\mathcal{W}) = \mathcal{L}_{cla}(\mathcal{X};\mathcal{W}_{d}) + \lambda\mathcal{L}_{seg}(\mathcal{X};\mathcal{W}) 
\label{total_loss}
\end{equation}
\subsection{ Overall Framework for pN-stage Classification}
The overall pipeline of our framework contains: (1)\textit{Data Preprocessing.} We firstly extract informative tissue regions from WSIs with Otsu algorithm \cite{otsu1979threshold}. Training patches and corresponding mask patches are augmented with random flipping, scaling, rotation, and cropping together. Color jittering and HSV augmentation are applied to training patches to overcome color variance. (2) \textit{Slide-level Metastasis Detection.} We extract ITC and boundary patches at first and add them to the original training set to train the full PFA-ScanNet. Only the detector of our PFA-ScanNet is used for inference. (3) \textit{Patient-level pN-stage Classification.} Morphological features (i.e., major axis length and metastasis area) are extracted from the probability maps to formulate feature vectors. We then utilize them to train a random forest classifier to classify the lymph nodes into four types, i.e., normal, ITC, Micro, and Macro. The patient's pN-stage is finally determined by the given rules in Chamelyon17 Challenge.
\section{Experimental Results}
\textbf{Datasets and Evaluation Metrics.} We evaluate our method on two public WSI datasets,  Camelyon16\footnote[1]{\url{http://camelyon16.grand-challenge.org/} } and Camelyon17\footnote[2]{\url{http://camelyon17.grand-challenge.org/} } datasets . The Camelyon16 dataset contains a total of 400 WSIs (270 training and 130 testing) with lesion-level annotations for all cancerous WSIs. The Camelyon17 dataset contains 1000 WSIs with 5 slides per patient (500 training and 500 testing), providing pN-stage labels for 100 patients in the training set and lesion-level annotations for only 50 WSIs where ITC and Micro have been included. For Camelyon17 Challenge, we use the whole Camelyon16 dataset and 215 WSIs including 50 slides with lesion-level annotations from Camelyon17 training set to train the network. We adopt two metrics provided in Camelyon16 Challenge to evaluate slide-level metastasis detection, including AUC and average FROC. The latter is an average sensitivity at 6 false positive rates: 1/4, 1/2, 1, 2, 4, and 8 per WSI. For pN-stage classification, we utilize quadratic weighted Cohen's kappa provided in Camelyon17 Challenge as the evaluation metric.\\
\\
\textbf{Implementation Details.} We implement our method using TensorFlow library on the workstation equipped with four NVIDIA  TITAN Xp GPUs.
The sizes of training patches and mask patches are $ 692 \times 692$ and $ 512\times512 $, respectively. Our model can take ROIs with a size up to $ 2708 \times 2708$ (determined by the memory capacity of GPU) for inference and outputs a $ 64 \times 64$ sized probability tile. To maximize the performance of synergistic learning, we set the truncated point $ \gamma $  in Equation (\ref{loss}) as 0.04. The hyperparameter $ \lambda $ is set to 0.5 in Equation (\ref{total_loss}). SGD optimizer is used to optimize the whole network with momentum of 0.9 and learning rate is initialized as 0.0001.\\
\\
\textbf{Quantitative Evaluation and Comparison.} We validate our method on Camelyon16 testing set and Camelyon17 testing set with ground truths held out. Results of Camelyon17 Challenge are provided by organizers. Table \ref{tab} compares our method with top-ranking teams as well as the state-of-the-art methods. It is observed that our method without synergistic learning (\textit{PFA-ScanNet w/o SL}) achieves striking improvements (3\% in FROC and 14\% in kappa score) compared 
\begin{table}
	\centering
	\caption{Comparison with different approaches on Camelyon16 and Camelyon17 Challenges. Runtime per ROI and per WSI are reported (unit: minute) in this table.}
	\label{tab}
	\begin{threeparttable}
	\resizebox{\textwidth}{!}{%
	\begin{tabular}{cccc|ccc}
			\hline
			\multicolumn{4}{c|}{Camelyon16 Challenge} & \multicolumn{3}{c}{Camelyon17 Challenge} \\ 
			Method & Runtime(ROI/WSI) & AUC & FROC & Team & Runtime(ROI/WSI) & Kappa Score \\ \hline
			Harvard \& MIT\cite{wang2016deep}$ ^{\ast} $  & 0.668/267.2 & \textbf{99.4\%} & 80.7\% & IMT Inc. (Fast ScanNet) & 0.020/8.0 & 0.778 \\
			NCRF\cite{li2018cancer} & 0.743/297.2 & - & 81.0\% & MIL-GPAT$ ^{\ast} $ & 0.247/98.8 & 0.857 \\
			Fast ScanNet\cite{fastscannet} & 0.020/8.0 & 98.7\% & 85.3\% & VCA-TUe & 1.162/464.8 & 0.873 \\
			Lunit Inc.\cite{lunit} & 1.136/454.4 & 98.5\% & 85.5\% & HMS-MGH-CCDS & 0.067/26.8 & 0.881 \\
			B.Kong et al.\cite{kongbin} & \textbf{0.014/5.6} & - & 85.6\% & ContextVision$ ^{\ast} $ & 1.636/654.4 & 0.883 \\
			LYNA \cite{liu2018artificial}$ ^{\ast} $& 1.155/462.0 & 99.3\% & 86.1\% & Lunit Inc. (2017 winner)$ ^{\ast} $ & 1.136/454.4 & 0.899 \\
			Y.Liu et al.\cite{liu2017detecting}$ ^{\ast} $ & 1.155/462.0 & 97.7\% & 88.5\% & DeepBio Inc.$ ^{\ast} $ & 1.583/633.2 & \textbf{0.957} \\ \hline\hline
			PFA-ScanNet w/o SL & 0.018/7.2 & 98.3\% & 87.8\% & PFA-ScanNet w/o SL & \textbf{0.018/7.2} & 0.887 \\
			\textbf{PFA-ScanNet} & 0.018/7.2 & 99.2\% & \textbf{89.1\%} &   \textbf{PFA-ScanNet} & \textbf{0.018/7.2} & 0.905 \\
			\textbf{PFA-ScanNet$ ^{\ast} $} & 0.018/7.2 & 98.8\% &\textbf{ 90.2\% }& -  & - & - \\ \hline
	\end{tabular}%
	}
	\begin{tablenotes}
		\footnotesize
		\item[ ] Note: $^{\ast}  $ denotes methods using model ensembles. 
	\end{tablenotes}
\end{threeparttable}
\end{table}
with previous Fast ScanNet, demonstrating the superiority of aggregating pyramidal features in scan-based models. 
After introducing synergistic learning, our PFA-ScanNet without model ensemble boosts the results to 89.1\% with regard to FROC (1st) and 99.2\% in terms of AUC on Camelyon16 testing set. It also achieves competitive kappa score of 0.905 on Camelyon17 testing set, surpassing the Challenge winner (Lunit Inc.) who utilized model ensembles. Furthermore, we trained another two models with dense coefficient $ \alpha =16, 32 $ for model ensembles, and the result can reach 90.2\% in terms of FROC, outperforming the state-of-the-art method by 2\%.

\begin{figure*}
	\centering
	\includegraphics[width=\textwidth]{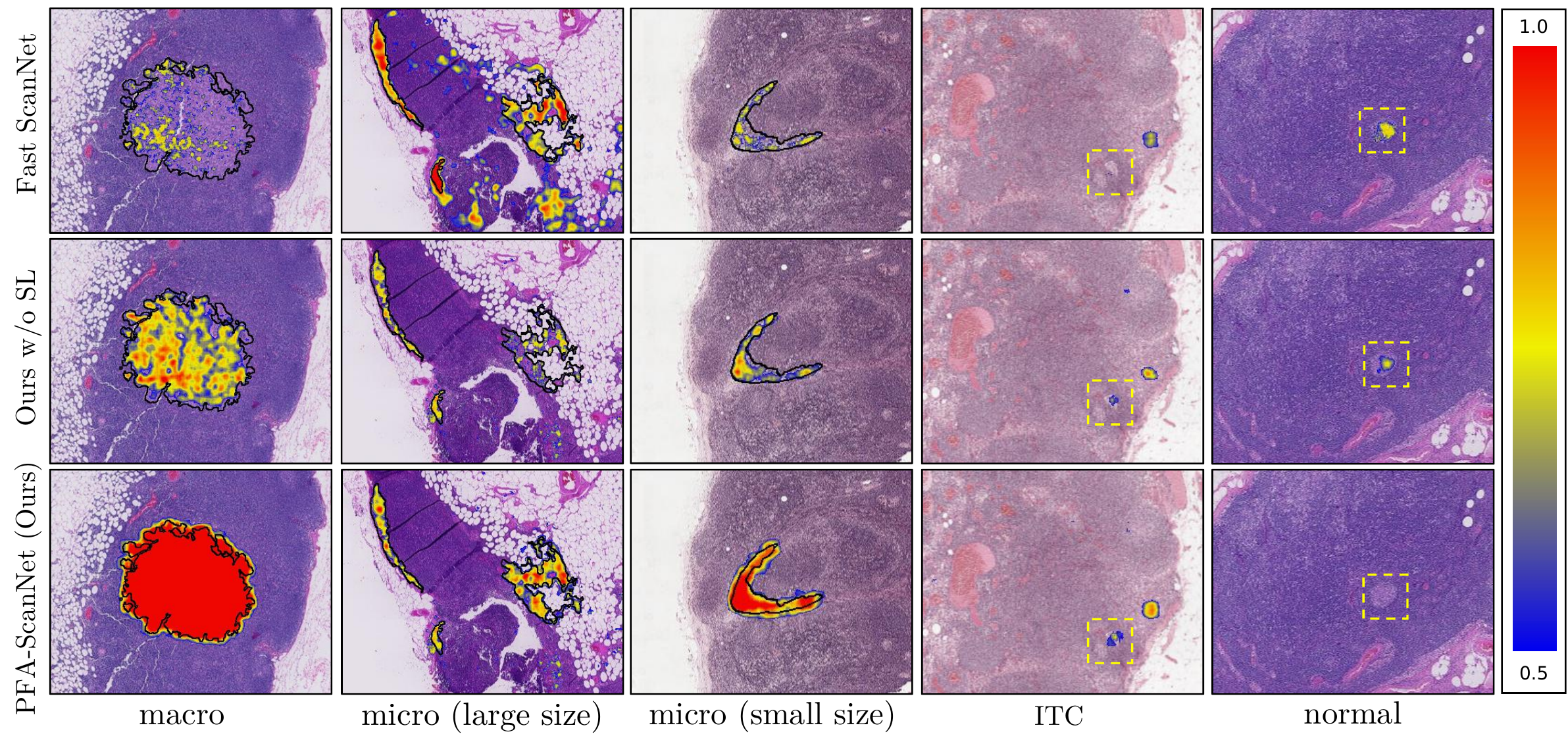}
	\caption{Typical examples of probability maps. The colors ranging from blue to red denote low to high probability. The lesion-level annotation is shown in black. }
	\label{fig:result}
\end{figure*}

For the speed performance, we measure the time cost of each method on a $ 2708\times2708 $ sized ROI with scanning stride of 32 (corresponding to the dense coefficient $ \alpha=16 $) using one single GPU. Since a typical WSI consists of around 400 ROIs in average with size $ 2708\times2708 $, the runtime can be converted from per ROI to per WSI. As illustrated in Table \ref{tab}, our method shows leading speed advantages over the state-of-the-art methods on Camelyon16 and Camelyon17 Challenges. Note that the proposed PFA-ScanNet is faster than Fast ScanNet in the inference phase (7.2 min vs. 8.0 min). Our large capacity network is even on par with B.Kong's method (a lightweight network) \cite{kongbin} in terms of speed performance while achieving notably higher accuracy on detection results. Besides, our method takes only 1.1\% time of the DeepBio Inc. to obtain probability maps (7.2 min vs. 633.2 min) and achieves competitive kappa score.
\\
\\
\textbf{Qualitative Analysis.} Fig. \ref{fig:result} visualizes metastasis detection results of five typical cases. 
As we can observe, the proposed PFA-ScanNet can generate more complete  results with higher probability for metastatic regions compared with the relatively sparse predictions from \textit{PFA-ScanNet w/o SL} and \textit{Fast ScanNet}. It thus increases the ability to detect macro- and micro-metastases, which is of great importance in clinical practice. 
Besides, the challenging ITC cases (see yellow boxes) can be detected by our method with few false positives, highlighting the advantage of PFA-ScanNet and proposed synergistic learning. 
\section{Conclusions}
Automatic cancer metastasis analysis is essential for cancer staging and following patient's treatment. In this paper, we propose the PFA-ScanNet with synergistic learning for metastasis detection and pN-stage classification to improve the accuracy close to clinical usage while maintaining the computational efficiency. Competitive results have been demonstrated on the Camelyon16 and Camelyon17 datasets with a much faster speed. Inherently our method can be applied to a wide range of medical image classification tasks to boost the analysis of gigapixel WSIs.
\subsubsection{Acknowledgments.}
This work was supported by 973 Program (Project No. 2015CB351706), Research Grants Council of Hong Kong Special Administrative Region (Project No. CUHK14225616), Hong Kong Innovation and Technology Fund (Project No. ITS/041/16), grants from the National Natural Science Foundation of China (Project No. U1613219).

%
%
%
%

\bibliographystyle{splncs04}
\bibliography{refs}

\end{document}